\documentclass[sigconf]{acmart}
\usepackage{fancyhdr}
\renewcommand\footnotetextcopyrightpermission[1]{} 
\AtBeginDocument{%
  \providecommand\BibTeX{{%
    \normalfont B\kern-0.5em{\scshape i\kern-0.25em b}\kern-0.8em\TeX}}}
\usepackage{graphicx}
\usepackage{float}
\usepackage{subfigure}
\usepackage{algorithm}
\usepackage{algpseudocode}
\usepackage{amsmath}

\begin{document}

\title{Instance-Aware Graph Convolutional Network for Multi-Label Classification}

\author{Yun Wang*, Tong Zhang*, Zhen Cui, Chunyan Xu, Jian Yang}
\authornote{Both authors contributed equally to this research.}

\affiliation{%
  \institution{PCA Lab, Key Lab of Intelligent Perception and Systems for High-Dimensional Information of Ministry of Education, and Jiangsu Key Lab of Image and Video Understanding for Social Security, 
  	\\School of Computer Science and Engineering, Nanjing University of Science and Technology, Nanjing, China}
  \city{\{yun.wang, tong.zhang, zhen.cui, cyx, csjyang\}@njust.edu.cn}
}


\begin{abstract}
  Graph convolutional neural network (GCN) has effectively boosted the multi-label image recognition task by introducing label dependencies based on statistical label co-occurrence of data. However,  in previous methods,  label correlation is computed based on statistical information of data and therefore the same for all samples, and this makes graph inference on labels insufficient to handle huge variations among numerous image instances. In this paper, we propose an instance-aware graph convolutional neural network (IA-GCN) framework for multi-label classification. As a whole, two fused branches of sub-networks are involved in the framework: a global  branch modeling the whole image and a region-based branch exploring dependencies among regions of interests (ROIs). For label diffusion of instance-awareness in graph convolution, rather than using the statistical label correlation alone, an image-dependent label correlation matrix (LCM), fusing both the statistical LCM and an individual one of each image instance, is constructed for graph inference on labels to inject adaptive information of label-awareness into the learned features of the model. Specifically, the individual LCM of each image is obtained by mining the label dependencies based on the scores of labels about detected ROIs. In this process, considering the contribution differences of ROIs to multi-label classification, variational inference is introduced to learn adaptive scaling factors for those ROIs by considering their complex distribution. Finally,  extensive experiments on MS-COCO and VOC datasets show that our proposed approach outperforms existing state-of-the-art methods.
\end{abstract}




\maketitle

\begin{figure}[h]
	\centering
	\includegraphics[width=7cm, height=8cm]{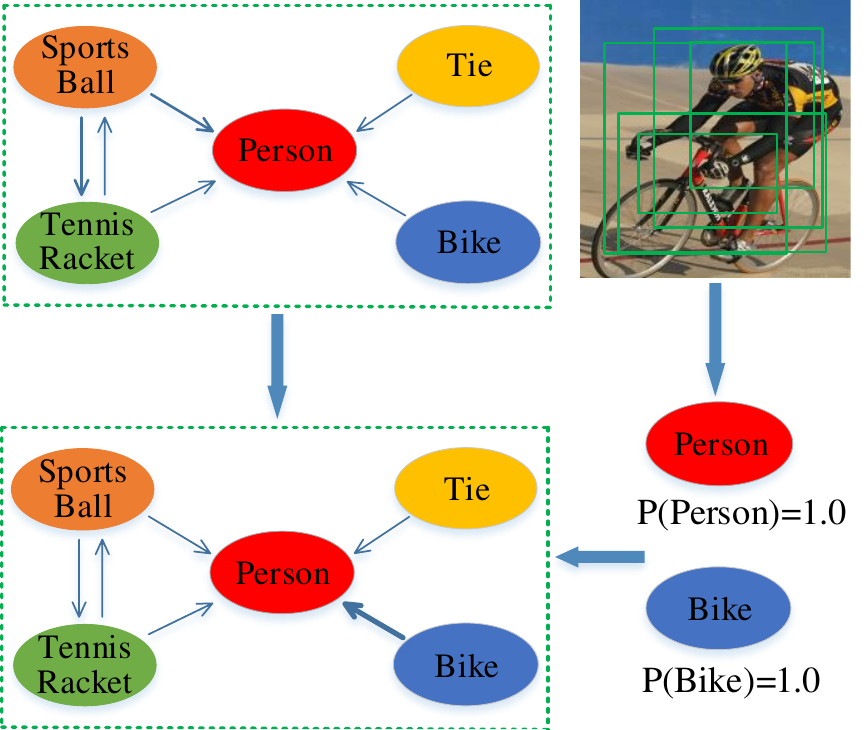}
	\caption{We first construct the directed graph according to the conditional probabilities of labels statistical co-occurrence information from training set, and then predict the labels scores of regions extracted from each current image. We apply the labels scores about regions to enhance their corresponding correlation in the directed graph, which means that the arrow lines among the predicted labels are bold.}
	\Description{The 1907 Franklin Model D roadster.}
	\label{P1}
\end{figure} 

\section{Introduction}

As a fundamental task in computer vision, multi-label image recognition aims to accurately and simultaneously recognize multiple objects present in an image. Compared to single-label image classification, multi-label recognition is more challenging because of  usually complex scene, more wide label space, and implicit correlation of objects. In view of the natural co-occurrence of objects in the real-world scene,  multi-label image classification is more practical than the single-label one, and has received wide attention \cite{Guillaumin2009TagPropDM,Boutell2004LearningMS,Li2016HumanAR,Ge2018ChestXC} in recent years.

Numerous algorithms have been proposed for multi-label image classification. In early works, deep Convolutional Neural Networks (CNNs) used for single-label recognition \cite{Szegedy2016RethinkingTI,He2016DeepRL,Huang2017DenselyCC,Simonyan2015VeryDC} are leveraged for the multi-label task by treating the multi-label recognition as a set of binary classification tasks. Although boosting  the accuracy of multi-label classification, however, this type of methods is still limited due to the ignorance of co-occurrence among objects, which can be reflected in label correlation. To model label dependencies, three lines of works have been proposed in recent literatures including attention mechanism based \cite{Szegedy2016RethinkingTI,Wang2017MultilabelIR}, recurrent neural network based \cite{Wang2016CNNRNNAU}, and graph based algorithms \cite{Li2016ConditionalGL,Li2014MultilabelIC,Chen2019MultiLabelIR}. Specifically, in graph based methods, graph convolution is introduced to characterize label correlations by diffusing label dependencies with a label correlation matrix (LCM). With graph inference on labels, graph convolutional neural network (GCN) and its variants \cite{Chen2019MultiLabelIR,Wang2020MultiLabelCW} have reported state-of-the-art performances in recent literatures. 

The success of GCN \cite{Kipf2017SemiSupervisedCW} indicates the significance of label correlation captured through the LCM for promoting the multi-label classification. In previous GCN \cite{Chen2019MultiLabelIR,Wang2020MultiLabelCW} based works, the LCM is global and dataset-dependent as it is obtained through the statistic of accessible data. However, in view of the large variation among images, the prior knowledge of label correlation may not well suit for all samples. For instance, the statistical co-occurrence of the bike and person is low, which may mislead the classification for the images of riders. Therefore, individual characteristics of label correlation should also be considered, and used to adaptively modify the prior knowledge. Here, we attempt to construct an adaptive individual LCM for each image instance by utilizing the rough classification scores of ROIs. As shown in Figure \ref{P1}, as an intuitive understanding, if all ROIs indicate high appearance probabilities of some labels, such as person and bike, then the correlation between person and bike in statistical LCM should be accordingly enhanced for this specific image. 

In this paper, we propose an instance-aware graph convolutional neural network (IA-GCN) framework for multi-label classification. The core idea is to adaptively construct one image-dependent label correlation matrix (ID-LCM) for each given image, which better favours the graph inference on labels. For framework construction, considering the previous success of GCN-based methods \cite{Chen2019MultiLabelIR,Wang2020MultiLabelCW}, we build two fused branches of sub-networks in the framework: a global branch modeling the whole image, and an additional region-based branch inferring on ROIs. Moreover, graph inference on labels is conducted to inject label-awareness into the both branches. In this process, different from previous works using statistical LCM \cite{Chen2019MultiLabelIR}, an image-dependent LCM is constructed by fusing both the statistical LCM and an individual one of each image instance. Specifically, the individual LCM of each image is obtained by mining the label dependencies based on the scores of detected ROIs. Considering the contribution differences of ROIs to multi-label classification, during the generation of the  individual LCM, variational inference \cite{Kingma2014AutoEncodingVB} is introduced to learn adaptive scaling factors for those ROIs by considering their complex distribution. As a result, the image-dependent LCM is flexible and benefits the proposed framework in adaptively propagating information on labels for each image instance. Finally, the learned features of both the global and local branches are fused and jointly modeled for the multi-label classification. We test  the performance on MS-COCO and VOC datasets, and the results show that our proposed approach outperforms existing state-of-the-art methods.

The main contributions of this paper are as follows:
\begin{itemize}
	\item We propose a novel IA-GCN framework for the multi-label classification task by jointly modeling the global context of the whole image and local dependencies of ROIs with adaptive information propagation on labels.
	\item A novel image-dependent LCM is constructed based on both the statistical LCM and an individual one of each image, which endows graph convolution flexibility to handle huge correlation variations among numerous image instances. 
	\item We introduce variational inference to explore label dependencies by considering the distribution of ROI appearances, which results in an adaptive LCM for each image instance.
	\item We report the state-of-the-art performances on both MS-COCO and VOC datasets, which verifies the effectiveness of our framework.
\end{itemize}

\section{Related work}
\begin{figure*}[!t] 
	\centering 
	\includegraphics[width=1\textwidth]{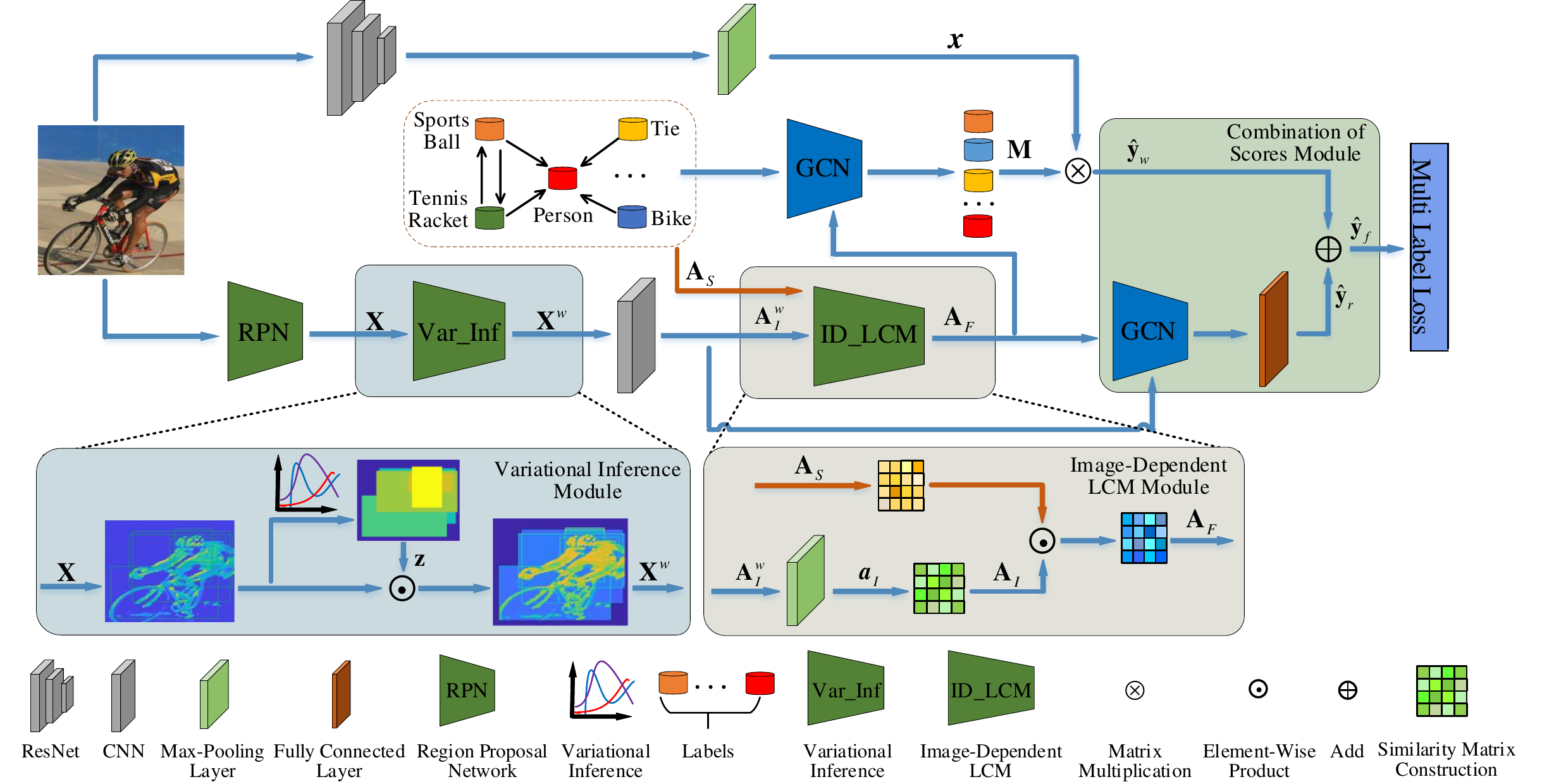}
	\caption{Overall framework of our IA-GCN model for multi-label image recognition. The whole architecture can be divided into a global branch and a region-based branch. In the global branch, we first model the image global feature and label dependencies by ResNet model and GCN respectively, and then obtain a set of labels scores $\hat{\mathbf{y}}_{_w}$ on the global branch by applying the learned label dependencies $\mathbf{M}$ to the image global feature $\boldsymbol{x}$. In the region-based branch, we extract a fixed number of ${N}$ ROIs by RPN, and then introduce variational inference to weight the ROIs by considering their complex distribution. The weighted regions $\mathbf{X}^{w}$ generate an individual LCM $\mathbf{A}_{_I}$ of the current image instance, which is used to fuse with the statistical LCM $\mathbf{A}_{_S}$. Simultaneously, we get another set of labels scores $\hat{\mathbf{y}}_{_r}$ on the region-based branch by exploring dependencies among weighted ROIs. Finally, we integrate two sets of labels scores as final labels scores $\hat{\mathbf{y}}_{_f}$ .} 
	\label{Pipeline} 
\end{figure*}

We first review the related works of multi-label image recognition, and then introduce variational inference and Graph Convolutional Neural Network. 

\textbf{Multi-Label Image Recognition. } An initial approach to dealing with multi-label recognition task was to divide it into multiple independent single-label tasks by training a binary classifier for each label. such as the BR method \cite{Tsoumakas2007MultiLabelCA} proposed by Tsoumakas. However, the performance of this method is limited by ignoring the correlation among labels. Many researchers proposed various methods for capturing label correlation and have achieved great success. Li et al. \cite{Li2014MultilabelIC} proposed to use probabilistic graph models for formulating the co-occurrence of labels. Gong et al. \cite{Gong2013DeepCR} discovered that using a weighted approximated-ranking loss function to train CNN could achieve better performance. Furthermore, Wang et al. \cite{Wang2016CNNRNNAU} combined CNN with RNN to learn a joint image-label embedding for characterizing the semantic label dependency and the image-label correlation. In addition, some researchers also applied attention mechanisms to capture label correlation. Wang et al. \cite{Wang2017MultilabelIR} utilized a spatial transformer layer to locate attentional regions and then used long short-term memory (LSTM) to obtain label correlation. Zhu et al. \cite{Zhu2017LearningSR} proposed to learn a spatial regularization network in order to explore label relevance.

\textbf{Variational Inference. } There are many problems that are difficult to find their exact solution. Thus, many researchers are committed to finding the approximate solutions of these problems. Variational inference \cite{Kingma2014AutoEncodingVB} is a common method to find the approximate solutions. Agakov \cite{Barber2003TheIA} explored variational bounds on mutual information without considering the objective of the information bottleneck. Mohamed and Rezende \cite{Mohamed2015VariationalIM} successfully applied variational inference to deep neural networks by exploring the variational boundaries of mutual information based on reinforcement learning. Chalk et al. \cite{Chalk2016RelevantSC} proposed to achieve nonlinear mapping by the kernel technique and obtained the variational lower bound of the information bottleneck objective. Alexander et al. \cite{Alemi2016DeepVI} proposed a Deep VIB model, which applied a neural network to parameterize the information bottleneck model and could obtain an approximate solution of the information bottleneck.

\textbf{Graph Convolutional Neural Network. } The graph is a more effective tool when we explore the correlation of object structure. Li et al. \cite{Li2014MultilabelIC} used the maximum spanning tree algorithm to create a tree-structured label graph. Lee et al. \cite{Lee2018MultilabelZL} utilized knowledge graphs to describe label dependency. Recently, since GCN \cite{Kipf2017SemiSupervisedCW} was proposed, it has achieved great success when it was applied to the modeling of non-grid structures. Chen et al. \cite{Chen2019MultiLabelIR} built a directed graph of labels and then used GCN to learn an inter-dependent label classifier. Furthermore, Wang et al. \cite{Wang2020MultiLabelCW} proposed to superimpose a knowledge prior label graph into a statistical label graph. In this paper, we propose an instance-aware GCN. In detail, firstly, we extract ROIs from each image by Region Proposal Network (RPN) and build an individual LCM for each current image based on the rough labels scores of ROIs. Secondly, we construct an image-dependent LCM fusing both the statistical LCM and an individual LCM of each image instance to adaptively guide information propagation among labels by GCN. Meanwhile, this fused LCM is also used to explore correlation among ROIs by another GCN.

\section{Approach}

In this part, we first overview the entire IA-GCN architecture, and then introduce four key modules in detail including image-dependent LCM construction, graph convolution on labels, Regions Features Construction and Loss Function.

\subsection{Overview}

The whole structure of the proposed IA-GCN is illustrated in Figure \ref{Pipeline}, the purpose of which is to accurately predict the labels of those jointly present objects from given $C$ labels. In the learning process, the input image passes through two branches of sub-networks, i.e. a global branch to model the whole image and a region-based branch capturing local dependencies among detected ROIs. For the global branch, considering the success in previous literatures \cite{Chen2019MultiLabelIR}, we employ ResNet-101 \cite{He2016DeepRL} to extract high-level features $\boldsymbol{x}$ to describe the context of the whole image. For the region-based branch, we first employ a RPN module \cite{Ren2015FasterRT} to generate a fixed number of $N$ ROIs and then extract the corresponding features $\mathbf{X}=\{{\mathbf{X}_{_1},\mathbf{X}_{_2},...,\mathbf{X}_{_N}}\}$. Considering the contribution differences of those ROIs for multi-label classification, a variational inference module \cite{Kingma2014AutoEncodingVB} is constructed to learn adaptive scaling factors $\mathbf{z}$ for them. As a result, salient regions for multi-label classification may be highlighted by weighting. Then, the weighted regions $\mathbf{X}^{w}=\{{\mathbf{X}_{_1}^{w},\mathbf{X}_{_2}^{w},...,\mathbf{X}_{_N}^{w}}\}$ endow adaptiveness to the constructed LCMs. In order to inject label-awareness into the two branches above, graph inference  is performed on labels with a constructed image-dependent LCM $\mathbf{A}_{_F}$. Specifically, the image-dependent LCM $\mathbf{A}_{_F}$ is constructed by fusing a statistical LCM $\mathbf{A}_{_S}$ and an individual LCM $\mathbf{A}_{_I}$ of each image instance. In detail, the statistical LCM $\mathbf{A}_{_S}$ is obtained based on the statistic of accessible training data, and the individual one $\mathbf{A}_{_I}$ is constructed based on the rough scores  $\mathbf{A}_{_I}^{w}$ of weighted ROIs. Finally, the two branches are fused to jointly predict the probabilities for those present objects. To optimize the whole framework, multi-label loss is calculated for back-propagation to jointly tune the parameters.  

\subsection{Image-Dependent LCM Construction}

Many exiting works propose to utilize the co-occurrence of labels or the conditional probabilities of labels \cite{Chen2019MultiLabelIR} which are both called the statistical information of labels in our paper to construct LCM. However, these methods may be insufficient to handle huge variations among numerous image instances when we just apply global statistical information to guide a single sample. In order to amend these variations, we also need to consider individual labels distribution of each instance. Thus, we integrate the two types labels information for the purpose of achieving their relative balance. In other words, we inject instance information into statistics to realize instance-awareness about labels. We formally present its details as follows. 

To achieve label diffusion of instance-awareness and better adaptively guide information propagation among labels, we construct an image-dependent LCM $\mathbf{A}_{_F}$, which is generated by dot product between $\mathbf{A}_{_I}$ and $\mathbf{A}_{_S}$.
\begin{equation}
\mathbf{A}_{_F} = \mathbf{A}_{_S}\odot \mathbf{A}_{I}\in \mathbb{R}^{{C\times C}}
\end{equation}
For the $\mathbf{A}_{_S}\in \mathbb{R}^{{C\times C}}$, we construct it by the method of ML-GCN \cite{Chen2019MultiLabelIR}. For the $\mathbf{A}_{_I}$, we construct it based on the features of weighted ROIs $\mathbf{X}^{w}=\{\mathbf{X}_{_1}^{w},\mathbf{X}_{_2}^{w},...,\mathbf{X}_{_N}^{w}\}\in\mathbb{R}^{N \times D},      \mathbf{X}_{_j}^{w}\in \mathbb{R}^{D}$. Specifically, firstly, in order to explore the distribution of labels for each region, we apply CNN layers to learn rough scores of labels. 
\begin{equation}
\mathbf{A}_{_I}^{w} = f_{_{CNN}}(\mathbf{X}^{w})\in \mathbb{R}^{{N\times C}}
\end{equation}
where $(\mathbf{A}_{_I}^{w})_{ij}$ is the score of the $i$-th region about $j$-th label. For one column denoted $(\mathbf{A}_{_I}^{w})_{\cdot j}$, it represents the probabilities of $N$ regions about the $j$-th label. Considering that the goal of multi-label recognition is to judge whether each label of interest exists. Thus, for $j$-th column of $\mathbf{A}_{_I}^{w}$, we select the maximum value of $(\mathbf{A}_{_I}^{w})_{\cdot j}$ as the probability of the current image instance about the $j$-th label. We operate the scores matrix $\mathbf{A}_{_I}^{w}$ by column-wise max pooling, which also overcomes over-fitting to some extent.
\begin{equation}
\boldsymbol{a}_{_I} = f_{_{MP}}(\mathbf{A}_{_I}^{w})\in \mathbb{R}^{C}
\end{equation}
where $\boldsymbol{a}_{_I}$ can be viewed as the rough classification scores of the image instance. We construct individual LCM of the image instance inspired by the statistical information of labels.
\begin{equation}
\mathbf{A}_{_I} = \boldsymbol{a}_{_I}(\boldsymbol{a}_{_I})^{T}\in \mathbb{R}^{{C\times C}}
\end{equation}

\subsection{Graph Convolution on Labels}
The essential idea of GCN is to update the features of nodes by propagating information among nodes based on an adjacent matrix denoted $\mathbf{A}$. In GCN, the features of each node is a mixture of itself and its neighbors from the previous layer. We follow the common operation \cite{Kipf2017SemiSupervisedCW}. Every GCN layer can be formulated as a non-linear function:   
\begin{equation}
\mathbf{H}^{(l+1)} = f(\hat{\mathbf{A}}\mathbf{H}^{(l)}\mathbf{W}^{(l)})
\end{equation}
where $\hat{\mathbf{A}}$ is the normalized version of adjacent matrix $\mathbf{A}$. $\mathbf{H}^{(l)}\in \mathbb{R}^{C\times D^{(l)}}$ is the all nodes features at the $l$-th layer. $\mathbf{W}^{(l)}\in \mathbb{R}^{D^{(l)}\times D^{(l+1)}}$ is a transformation matrix and is learned in the training phase. $f(\cdot)$ is a non-linear activation function. GCN can capture deep features of nodes by stacking multiple GCN layers. 

We view each label as a node and infer its final features by GCN. We first use the fused LCM $\mathbf{A}_{_F}$ as adjacent matrix $\mathbf{A}$, and then utilize Glove as the initial labels representations $\mathbf{H}^{(0)}$ which serve as the inputs of GCN. Finally, we can obtain labels representations $\mathbf{M}\in \mathbb{R}^{C\times D}$ as object classifiers which are both inter-dependent and image-dependent via stacking multiple GCN layers. We apply the classifiers to the global image features $\boldsymbol{x}\in \mathbb{R}^{D}$ which comes from the global branch and then get a set of scores on all labels of the image instance $\hat{\mathbf{y}}_{_w}$. 
\begin{equation}
\hat{\mathbf{y}}_{_w} = \mathbf{M}\boldsymbol{x}\in \mathbb{R}^{C}
\end{equation}

\subsection{Region Feature Construction}
\begin{table*}
	\caption{Performance comparisons with state-of-the-art methods on the MS-COCO dataset. }
	\label{tab:commands}
	\begin{tabular}{cccccccc}
		\toprule
		Methods &mAP & CP& CR& CF1& OP& OR& OF1\\
		\midrule
		CNN-RNN \cite{Wang2016CNNRNNAU} & 61.2&-&-&-&-&-&-\\
		SRN \cite{Zhu2017LearningSR} &77.1&81.6&65.4&71.2&82.7&69.9&75.8\\
		ResNet-101 \cite{He2016DeepRL}&77.3&80.2&66.7&72.8&83.9&70.8&76.8\\
		Multi-Evidence \cite{Ge2018MultievidenceFA} &-&80.4&70.2&74.9&85.2&72.5&78.4\\
		ML-GCN \cite{Chen2019MultiLabelIR}&82.9&83.7&72.7&77.9&84.5&76.2&80.1\\
		KSSNet \cite{Wang2020MultiLabelCW}&83.7&84.6&73.2&77.2&\textbf{87.8}&76.2&81.5\\
		\midrule
		IA-GCN & \textbf{86.3}&\textbf{85.1}&\textbf{77.1}&\textbf{80.9}&85.5&\textbf{80.5}&\textbf{82.9}\\
		\bottomrule
	\end{tabular}
\end{table*}

\begin{table*}
	\caption{Comparisons of AP and mAP with state-of-the-arts methods on the VOC dataset. }
	\renewcommand\tabcolsep{1.5pt}
	\label{tab:commands}
	\begin{tabular}{cccccccccccccccccccccc}
		\toprule
		Methods &aero & bike& bird& boat& bottle& bus& car&cat&chair&cow&table&dog&horse&motor&person&plant&sheep&sofa&train&tv&mAP\\
		\midrule
		CNN-RNN \cite{Wang2016CNNRNNAU}&96.7&83.1&94.2&92.8&61.2&82.1&89.1&94.2&64.2&83.6&70.0&92.4&91.7&84.2&93.7&59.8&93.2&75.3&99.&78.6&84.0\\
		RLSD \cite{Zhang2018MultilabelIC}&96.4&92.7&93.8&94.1&71.2&92.5&94.2&95.7&74.3&90.0&74.2&95.4&96.2&92.1&97.9&66.9&93.5&73.7&97.5&87.6&88.5\\
		VeryDeep \cite{Simonyan2015VeryDC}&98.9&95.0&96.8&95.4&69.7&90.4&93.5&96.0&74.2&86.6&87.8&96.0&96.3&93.1&97.2&70.0&92.1&80.3&98.1&87.0&89.7\\
		ResNet-101 \cite{He2016DeepRL}&\textbf{99.5}&97.7&97.8&96.4&65.7&91.8&96.1&97.6&74.2&80.9&85.0&\textbf{98.4}&96.5&95.9&98.4&70.1&88.3&80.2&\textbf{98.9}&89.2&89.9\\
		FeV+LV \cite{Yang2016ExploitBB}&97.9&97.0&96.6&94.6&73.6&93.9&96.5&95.5&73.7&90.3&82.8&95.4&97.7&95.9&98.6&77.6&88.7&78.0&98.3&89.0&90.6\\
		HCP \cite{Wei2016HCPAF}&98.6&97.1&98.0&95.6&75.3&94.7&95.8&97.3&73.1&90.2&80.0&97.3&96.1&94.9&96.3&78.3&94.7&76.2&97.9&91.5&90.9\\
		RNN-Attention \cite{Wang2017MultilabelIR}&98.6&97.4&96.3&96.2&75.2&92.4&96.5&97.1&76.5&92.0&87.7&96.8&97.5&93.9&98.5&81.6&93.7&82.8&98.6&89.3&91.9\\
		Atten-Reinforce \cite{Chen2018RecurrentAR}&98.6&97.1&97.1&95.5&75.6&92.8&96.8&97.3&78.3&92.2&87.6&96.9&96.5&93.6&98.5&81.6&93.1&83.2&98.5&89.3&9.2\\
		ML-GCN \cite{Chen2019MultiLabelIR}&98.9&97.5&97.1&97.4&79.4&94.1&96.9&97.1&81.9&93.0&84.2&96.8&97.4&95.5&98.7&84.5&96.4&82.7&98.5&91.3&93.0\\
		\midrule
		IA-GCN&99.4&\textbf{99.1}&\textbf{98.5}&\textbf{98.3}&\textbf{83.2}&\textbf{96.3}&\textbf{98.2}&\textbf{98.1}&\textbf{84.3}&\textbf{95.3}&\textbf{88.0}&98.0&\textbf{98.3}&\textbf{96.5}&\textbf{99.3}&\textbf{87.5}&\textbf{97.2}&\textbf{87.2}&98.6&\textbf{94.5}&\textbf{94.8}\\
		\bottomrule
	\end{tabular}
\end{table*}

We extract $N$ ROIs $\mathbf{X}=\{{\mathbf{X}_{_1},\mathbf{X}_{_2},...,\mathbf{X}_{_N}}\}\in\mathbb{R}^{N \times D},   \mathbf{X}_{_j}\in \mathbb{R}^{D}$ from each image by RPN. For the ROIs $\mathbf{X}$, considering that the regions may not be complete, some regions may contain useful objects while some regions may be noise. Thus, the contribution of ROIs to multi-label classification may be different. In order to explore the importance of different ROIs, we learn adaptive scaling factors for weighting these ROIs by considering their complex distribution.
\begin{equation}
\mathbf{X}^{w} = \mathbf{z}\odot \mathbf{X}\in \mathbb{R}^{N\times D}
\end{equation}
where $\mathbf{z}\in \mathbb{R}^{N}$ is the adaptive factor learned by variational inference. Specifically, we encode the mean parameters $\boldsymbol{w}_{{\mu}}\in \mathbb{R}^{D}$ and variance parameters $\boldsymbol{w}_{{\sigma^{2}}}\in \mathbb{R}^{D}$ of ROIs.

\begin{equation}
\mathbf{z}_{{\mu}} = \mathbf{X}\boldsymbol{w}_{{\mu}}\in \mathbb{R}^{N}
\end{equation}
\begin{equation}
\mathbf{z}_{{\sigma^{2}}} = \mathbf{X}\boldsymbol{w}_{\sigma^{2}}\in \mathbb{R}^{N}
\end{equation}
where $\mathbf{z}_{{\mu}} $ and $\mathbf{z}_{\sigma^{2}}$ are the learned means and variances. We sample one 
$\mathbf{z}\in \mathbb{R}^{N}$ from $\mathit{N}(\mathbf{z}_{{\mu}},\mathbf{z}_{\sigma^{2}})$, which is a distribution with $\mathbf{z}_{\mu}$ as the mean and $\mathbf{z}_{\sigma^{2}}$ as the variance. We also should note that our sampling operation uses "Reparameterization Trick" in order to allow variational inference to back-propagation. $N$ values of $\mathbf{z}$ are the weights of $N$ ROIs respectively.

From the Eq.(2), we can obtain the $\mathbf{A}_{_I}^{w}$ which contains the scores of all regions about all labels. We think $\mathbf{A}_{_I}^{w}$ is also the feature of ROIs in label space to some extent. However, due to the simplicity of its process, the accuracy of $\mathbf{A}_{_I}^{w}$ may be rough. What's worse, it doesn't consider the correlation of labels. Thus, we view each region as a node and apply GCN again to explore the correlations among the ROIs. We also use the $\mathbf{A}_{_F}$ as the adjacent matrix and use the  $\mathbf{A}_{_I}^{w}$ as the initial nodes representations. According to the above introduction about GCN, we can formulate the corresponding function. 
\begin{equation}
\hat{\mathbf{y}}_{_r} = f_{_{FC}}(f_{_{GCN}}(\hat{\mathbf{A}}_{_F}\mathbf{A}_{_I}^{w}\mathbf{W}_{_I}))\in \mathbb{R}^{C}
\end{equation}
We obtain another set of labels scores  $\hat{\mathbf{y}}_{_r}$ of the image instance on the region-based branch, which is generated according to the output of the fully connected layer after GCN. 

Based on the above discussion, two types of labels scores $\hat{\mathbf{y}}_{_w}$ and $\hat{\mathbf{y}}_{_r}$ have been acquired. They come from different branches and focus on different perspectives, which may be complementary. We merge the two sets for making full use of their effective information.
\begin{equation}
\hat{\mathbf{y}}_{_f} =\lambda\hat{\mathbf{y}}_{_w}+(1-\lambda)\hat{\mathbf{y}}_{_r}\in \mathbb{R}^{C}
\end{equation}
where $\lambda\in [\mathrm{0},\mathrm{1}]$ is a weight coefficient.

\subsection{Loss Function}
In order to make our model have good performance, we need to enable not only the recognition ability for labels, but also the adaptive generation ability of variational inference module. So our loss function consists of two parts.
\begin{equation}
\mathcal{L} = \mathcal{L}_{_{ML}}+ \mathcal{L}_{_{KL}}
\end{equation}
For the $\mathcal{L}_{_{ML}}$, it is the traditional multi-label loss function for multi-label recognition task.
\begin{equation}
 \mathcal{L}_{_{ML}} =-\frac{1}{C}\sum_{i=1}^{C} (\mathbf{y})_{i}log(\sigma(\hat{\mathbf{y}}_{_f})_{i})+(1-(\mathbf{y})_{i})log(1-\sigma(\hat{\mathbf{y}}_{_f})_{i})
\end{equation}
where $\mathbf{y}\in \mathbb{R}^{C}$ is the ground truth label of an image and $\mathbf{y}_{_i}\in{\{\mathrm{0},\mathrm{1}\}}$ denotes whether $i$-th label exits or not. $\sigma(\cdot)$ is the sigmoid function. For the $\mathcal{L}_{_{KL}}$, we construct it by Kullback–Leibler ($KL$) divergence.
\begin{equation}
\mathcal{L}_{_{KL}} =\frac{1}{N} \sum_{i=1}^{N} \mathit{KL(\mathit{N}((\mathbf{z}_{{\mu}})_{i},(\mathbf{z}_{\sigma^{2}})_{i})||\mathit{N}(\mathrm{0},\mathrm{1}))}
\end{equation}

\section{Experiments}

In this section, we first describe the datasets and evaluation metrics. Then, we report the implementation details and comparisons with state-of-the-art methods. Finally, we carry out ablation studies to evaluate the effectiveness of our modules.

\subsection{Datasets}
Two public datasets, MS-COCO \cite{Lin2014MicrosoftCC} and PASCAL VOC \cite{Everingham2009ThePV}, are used to test our model and other state-of-the-art methods.

MS-COCO dataset \cite{Lin2014MicrosoftCC} is a widely used dataset, which can be used for multi-label recognition, object detection, etc. It contains 82,081 training images and 40,504 validation images. The dataset covers 80 classes. Each image contains 2.9 labels on average. Due to the lack of ground truth labels on the test set, we evaluate the performance of all the methods on the validation set. 

PASCAL VOC dataset \cite{Everingham2009ThePV} is also a popular dataset which is used for multi-label recognition task. Its $trainval$ set contains 5,011 images and test set contains 4,952 images. 20 classes are involved in the dataset. The $trainval$ set is used to train our model and the test set is used to evaluate the performance.

\subsection{Evaluation Metrics}

\begin{table*}
	\caption{Performance comparisons between different modules on MS-COCO dataset. }
	\label{tab:commands}
	\begin{tabular}{cccccccc}
		\toprule
		Methods &mAP & CP& CR& CF1& OP& OR& OF1\\
		\midrule
		Base &82.9&83.7&72.7&77.9&84.5&76.2&80.1\\
		Base+ID\_LCM &85.4&\textbf{86.1}&73.3&79.2&\textbf{88.0}&75.6&81.3\\
		Base+ID\_LCM+Var\_Inf &86.0&84.7&77.0&80.6&85.3&80.1&82.6\\
		Base+ID\_LCM+Var\_Inf+Com\_Sco &\textbf{86.3}&85.1&\textbf{77.1}&\textbf{80.9}&85.5&\textbf{80.5}&\textbf{82.9}\\
		\bottomrule
	\end{tabular}
\end{table*}

\begin{table*}
	\caption{Performance comparisons between different modules on VOC dataset. }
	\label{tab:commands}
	\begin{tabular}{cccccccc}
		\toprule
		Methods &mAP & CP& CR& CF1& OP& OR& OF1\\
		\midrule
		Base &93.0&85.9&86.1&86.0&84.7&89.3&86.9 \\
		Base+ID\_LCM &94.0&85.5&89.7&87.6&87.1&90.0&88.5\\
		Base+ID\_LCM+Var\_Inf &94.4&84.5&\textbf{91.2}&87.7&86.2&92.2&89.1\\
		Base+ID\_LCM+Var\_Inf+Com\_Sco & \textbf{94.8}&\textbf{85.9}&91.0&\textbf{88.3}&\textbf{87.7}&\textbf{92.2}&\textbf{89.9}\\
		\bottomrule
	\end{tabular}
\end{table*}

In order to evaluate performance comprehensively and compare with other methods conveniently, we report the average per-class (CP), recall (CR), F1 (CF1), the average overall precision (OP), overall recall (OR), overall F1 (OF1) and the mean average precision (mAP). mAP, OF1 and CF1 are relatively more important among all evaluation metrics. Precision represents the proportion of true positive samples in all predicted positive samples. Recall indicates the proportion of all positive samples that are predicted to be positive. F1 is generally used to measure the comprehensive performance classifiers. 
\begin{equation}
\mathrm{OP} =\frac{\sum_{i=1}^{_C}{N}_{i}^{cor}}{\sum_{i=1}^{_C}{N}_{i}^{pre}} \quad \quad \quad \quad \mathrm{CP} =\frac{1}{C}\sum_{i=1}^{C}\frac{{N}_{i}^{cor}}{{N}_{i}^{pre}}
\end{equation}
\begin{equation}
\mathrm{OR} =\frac{\sum_{i=1}^{_C}{N}_{i}^{cor}}{\sum_{i=1}^{_C}{N}_{i}^{gt}} \quad \quad \quad \quad \mathrm{CR} =\frac{1}{C}\sum_{i=1}^{C}\frac{{N}_{i}^{cor}}{{N}_{i}^{gt}}
\end{equation}
\begin{equation}
\mathrm{OF1} =\frac{2 \times \mathrm{OP} \times \mathrm{OR} }{\mathrm{OP} + \mathrm{OR}} \quad \quad \quad \quad \mathrm{CF1} =\frac{2 \times \mathrm{CP} \times \mathrm{CR} }{\mathrm{CP} + \mathrm{CR}}
\end{equation}
where $C$ is the number of labels. ${N}_{i}^{cor}$ is the number of images that are correctly predicted for the $i$-th label. ${N}_{i}^{pre}$ is the number of predicted images for the $i$-th label. ${N}_{i}^{gt}$ is the number of ground truth images for the $i$-th label.

\subsection{Implementation Details}

\textbf{\quad Pre-processing. } The ResNet-101 \cite{He2016DeepRL} used to extract images global features is pre-trained on ImageNet \cite{Deng2009ImageNetAL}. For each image, we randomly crop and resize the input images into $448\times 448$. We adopt 300-dimensional Glove \cite{Pennington2014GloveGV} model to obtain the representations of labels as the initial label embeddings. If one label has multiple words, we average all embeddings of words on the same dimension as its overall representations.

\textbf{IA-GCN Details. } The GCN exploring the label correlations has 3 layers and its output dimensions are 512, 1024 and 2048 respectively. For the GCN capturing the correlations among ROIs, it contains 4 layers with the output dimensions of 256, 512, 1024, 2048 and a fully connected layer with the output dimension of $C$. All non-linear activation functions are all ReLU. The number of ROIs $N$ is 40. We set $\lambda$ in Eq.(11) to be 0.8. 

\textbf{Training Strategy. } During the training phase, SGD is used as the optimizer. Its momentum and weight decay are 0.9 and $10^{-4}$ respectively. Considering that the parameters of ResNet-101 have been pre-trained, in order to maintain the consistency of optimization degree of different parameters, we adopt different learning rate for them. The initial learning rate of ResNet module is 0.001 and the others is 0.01. All parameters decay by a factor of 10 for every 30 epochs. The network is trained for 120 epochs in total.

\subsection{Comparisons with State-of-the-Arts}

We compare our model with state-of-the-art methods on MS-COCO dataset, including CNN-RNN \cite{Wang2016CNNRNNAU}, SRN \cite{Zhu2017LearningSR}, ResNet-101 \cite{He2016DeepRL}, Multi-Evidence \cite{Ge2018MultievidenceFA}, ML-GCM \cite{Chen2019MultiLabelIR} and KSSNet \cite{Wang2020MultiLabelCW}. The specific results are presented in Table 1. The performance of KSSNet \cite{Wang2020MultiLabelCW} is best currently. It is a GCN+CNN model that captures the correlations among labels by superimposing the knowledge graph into the statistical graph. We can observe from Table 1 that our model outperforms KSSNet at almost all evaluation matrices. Specifically, our model obtains 86.3\% on mAP and outperforms KSSNet by 3.1\%. CF1 is improved from 77.2\% to 80.9\%. OF1 is also increased by 1.7\%. These improvements demonstrate the superiority of our model. What's more, compared with the baseline model ML-GCN \cite{Chen2019MultiLabelIR}, which only uses the statistical graph to model the correlations among labels, our performance outperforms ML-GCN \cite{Chen2019MultiLabelIR} at all evaluation metrics, which sufficiently demonstrates the effectiveness of our model.

Compared with state-of-arts methods on VOC dataset, including CNN-RNN \cite{Wang2016CNNRNNAU}, RSLD \cite{Zhang2018MultilabelIC}, VeryDeep \cite{Simonyan2015VeryDC}, ResNet-101 \cite{He2016DeepRL}, FeV+LV \cite{Yang2016ExploitBB}, HCP \cite{Wei2016HCPAF}, RNN-Attention \cite{Wang2017MultilabelIR}, Atten-Reinforce \cite{Chen2018RecurrentAR} and ML-GCN \cite{Chen2019MultiLabelIR}, our model still outperforms them. Quantitative results are reported in Table 2. ML-GCN \cite{Chen2019MultiLabelIR} is the current state-of-the-arts on VOC dataset. Compared with ML-GCN \cite{Chen2019MultiLabelIR}, our model obtain 94.8\% at mAP metric and outperforms by 1.9\%. Furthermore, our model can obtain higher results at average precision evaluation matrices of almost all labels. These improvements demonstrate the effectiveness of our model again.

\subsection{Ablation Studies}

\begin{figure*}[!htb]
	\centering 
	\subfigure[raw image $i$]{
		\centering
		\label{Segments.sub.1}
		\includegraphics[width=2.7cm,height=3cm]{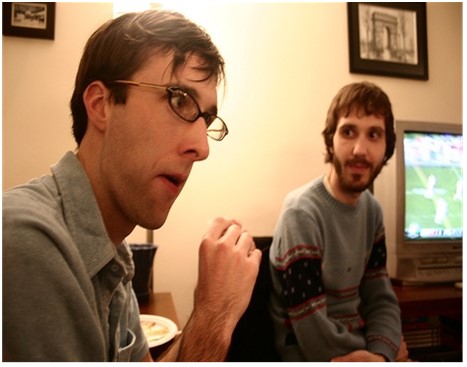}}
	\subfigure[detected regoins of $i$]{
		\centering
		\label{Segments.sub.2}
		\includegraphics[width=2.7cm,height=3cm]{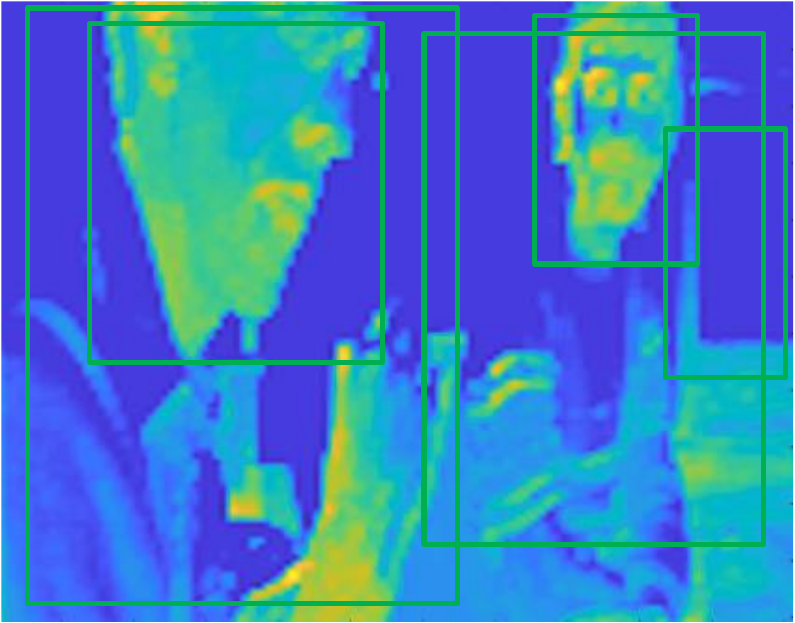}}
	\subfigure[weighted regions of $i$]{
		\centering
		\label{Segments.sub.3}
		\includegraphics[width=2.7cm,height=3cm]{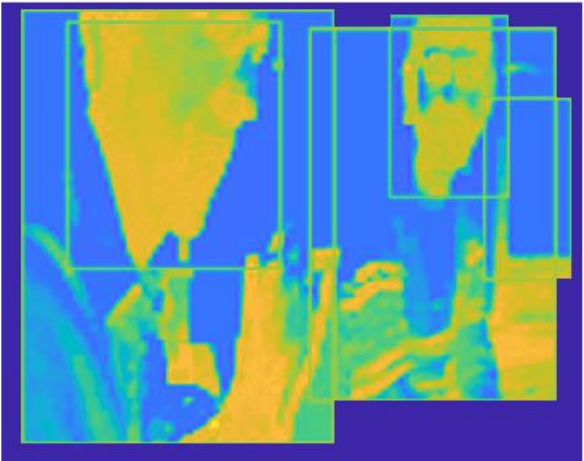}}
	\subfigure[raw image $j$]{
		\centering
		\label{Segments.sub.4}
		\includegraphics[width=2.7cm,height=3cm]{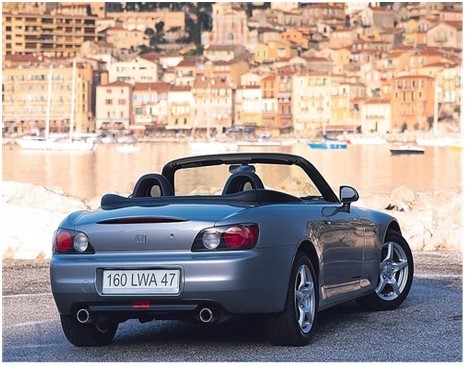}}
	\subfigure[detected regoins of $j$]{
		\centering
		\label{Segments.sub.5}
		\includegraphics[width=2.7cm,height=3cm]{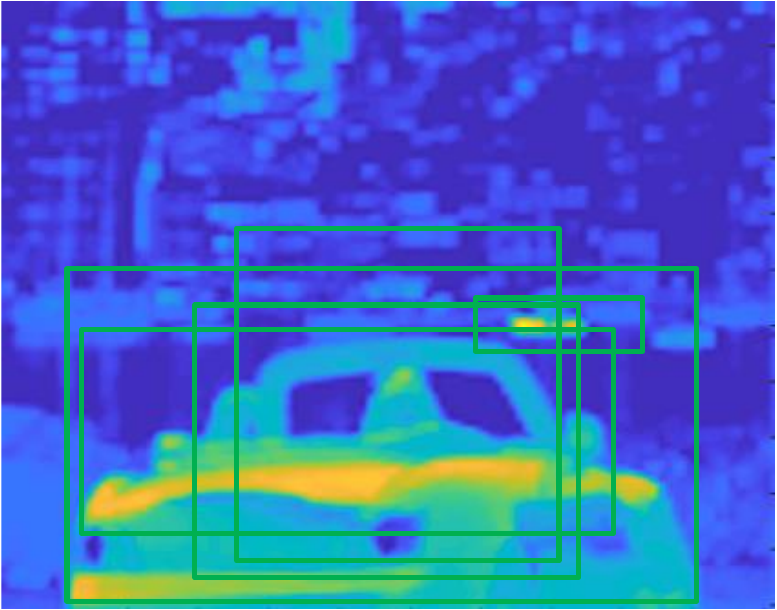}}
	\subfigure[weighted regions of $j$]{
		\centering
		\label{Segments.sub.6}
		\includegraphics[width=2.7cm,height=3cm]{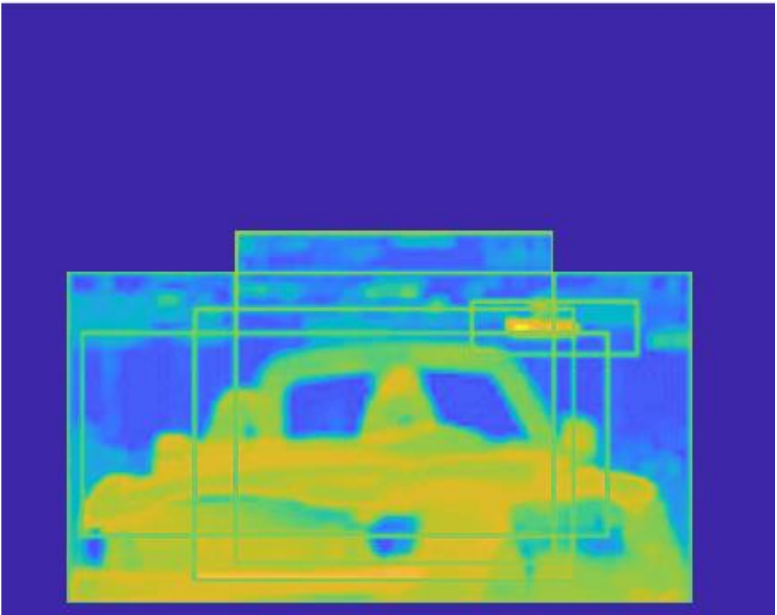}}
	\subfigure[raw image $m$]{
		\centering
		\label{Segments.sub.7}
		\includegraphics[width=2.7cm,height=3cm]{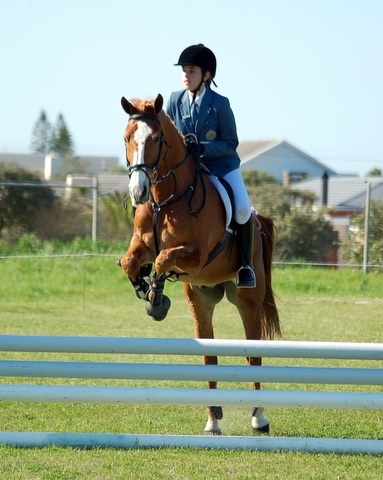}}
	\subfigure[detected regoins of $m$]{
		\centering
		\label{Segments.sub.8}
		\includegraphics[width=2.7cm,height=3cm]{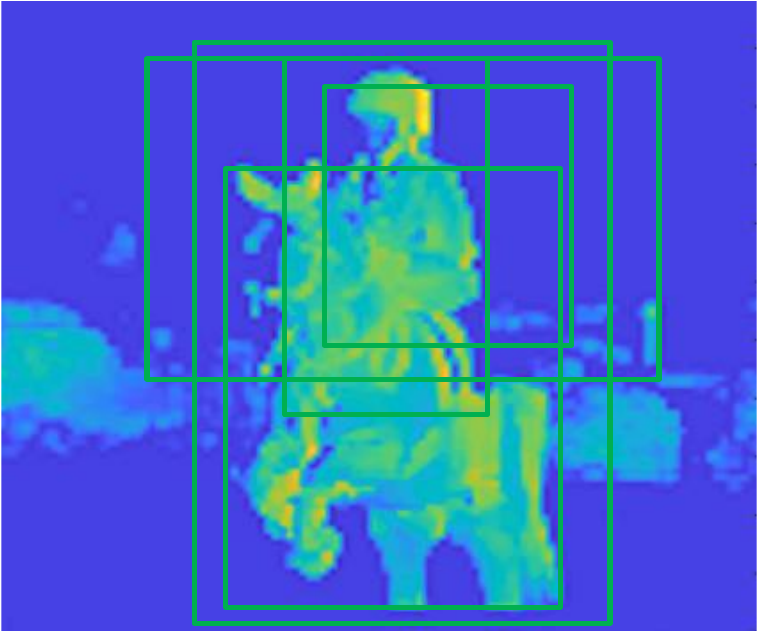}}
	\subfigure[weighted regions of $m$]{
		\centering
		\label{Segments.sub.9}
		\includegraphics[width=2.7cm,height=3cm]{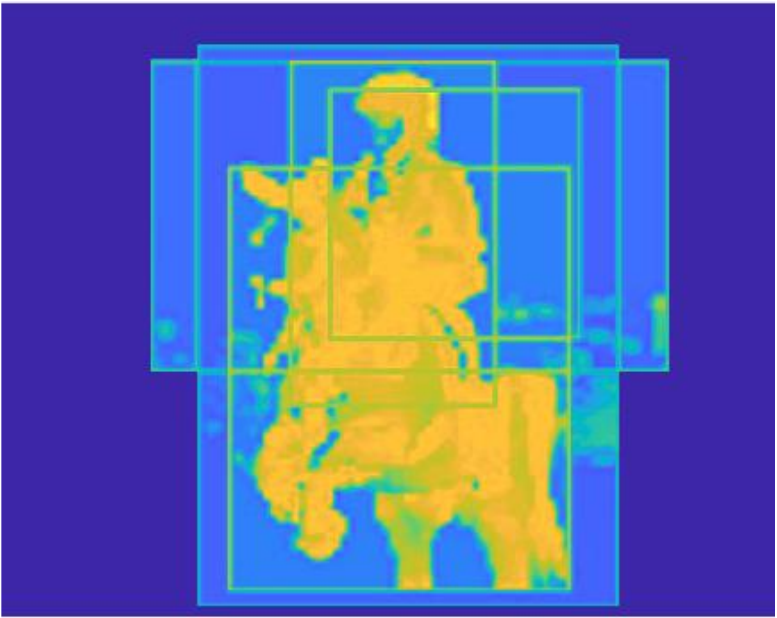}}
	\subfigure[raw image $n$]{
		\centering
		\label{Segments.sub.10}
		\includegraphics[width=2.7cm,height=3cm]{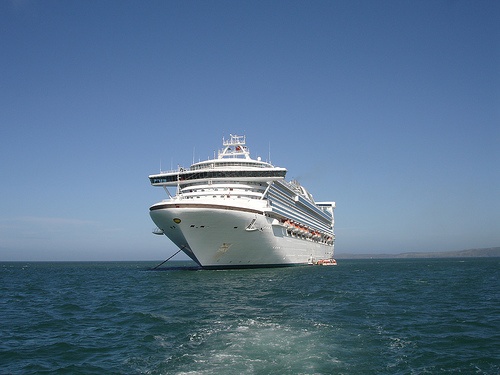}}
	\subfigure[detected regoins of $n$]{
		\centering
		\label{Segments.sub.11}
		\includegraphics[width=2.7cm,height=3cm]{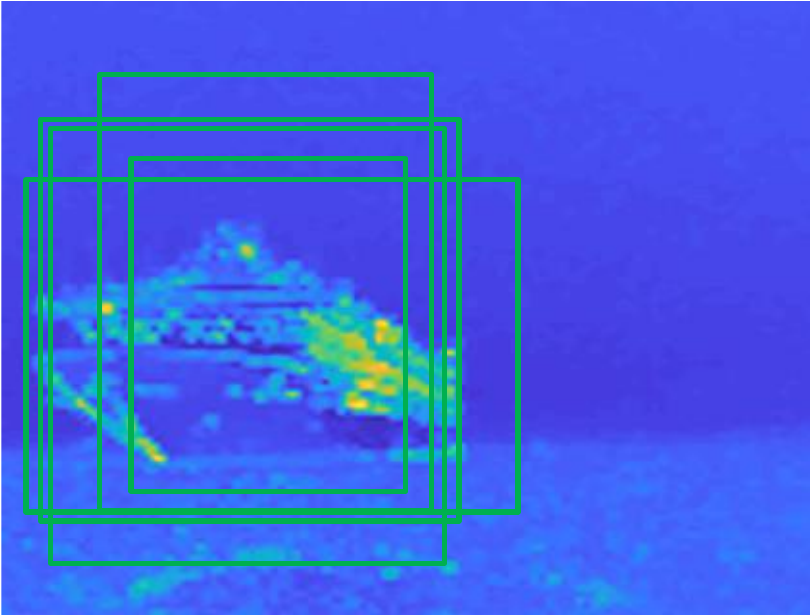}}
	\subfigure[weighted regions of $n$]{
		\centering
		\label{Segments.sub.12}
		\includegraphics[width=2.7cm,height=3cm]{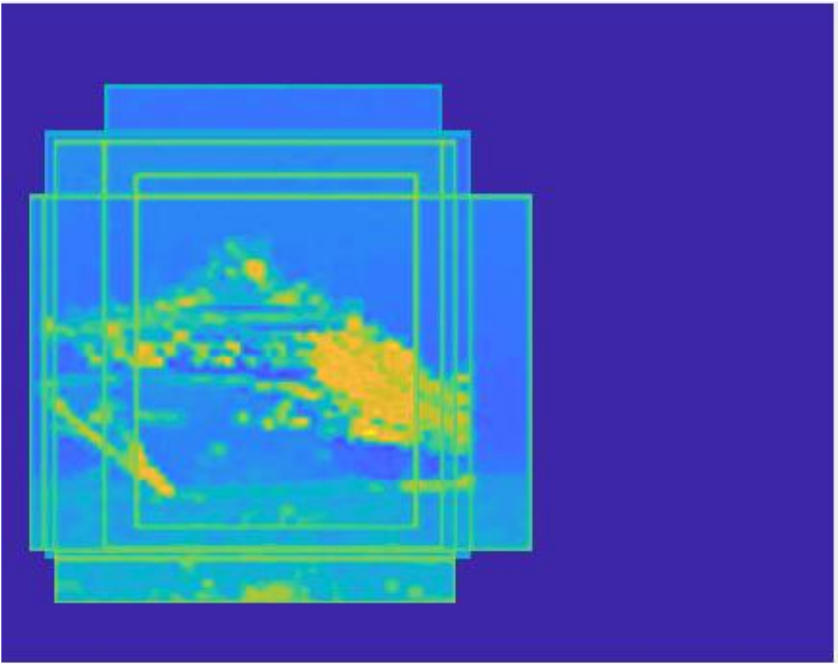}}
	\caption{Comparison of detection regions and weighted regions. The Var\_Inf module can effectively weight the detection regions.}
	\label{var} 
\end{figure*}

In this section, we analyze the effectiveness of each module in our model, including the image-dependent LCM, variational inference and the combination of two sets of scores. For the convenience of representation, we abbreviate each module into ID\_LCM, Var\_Inf and Com\_Sco respectively. Simultaneously, we denote ML-GCN \cite{Chen2019MultiLabelIR} as Base, which is the baseline model in our paper. From Table 3 and Table 4, we can observe that the performance of most of the indicators will be improved when we add more modules, especially that mAP, CF1 and OF1 indicators show a gradually increasing trend. 

\textbf{Effectiveness of ID\_LCM }
From Table 3 and Table 4, we can observe the comparison results between Base and Base+ID\_LCM on two datasets. The performance of most of the indicators is improved, which is caused by ID\_LCM. In the baseline, it only uses statistical information to construct the correlations of labels, However, statistical information comes from training set and it ignores the difference between the whole and the individual. Thus, it may be insufficient to handle huge variations among numerous image instances. Our ID\_LCM module considers both statistical information and individual label distribution. For each image instance, we construct an individual classifier, which contains the information of label distribution belonging to the instance, to enhance the label-awareness for the instance.

\textbf{Effectiveness of Var\_Inf }
We explore the effectiveness of the Var\_Inf module by adding the module to Base+ID\_LCM. It can be observed from Table 3 and Table 4 that the Var\_Inf module also improves the performance. For ROIs extracted from each image, Base+ID\_LCM thinks that all ROIs have the same importance and the contribution to multi-label classification is the same. However, because some regions may contain useful objects and some regions may be noise, if we treat them equally, those useless noise information will disturb our final classification results. Thus, we need to enhance those useful areas and suppress those noise areas. As shown in Figure \ref{var}, in this process, some of detected regions contains target objects, and some detect background information. However, after the regions are processed by this Var\_Inf module, the features of regions containing target objects are enhanced and their colors become more prominent and striking.  By considering their complex distribution, we apply the Var\_Inf module to learn adaptive scaling factors that are used to weight ROIs. For weighted ROIs, more useful information will be sent to the later network, and that useless information will be blocked.

\begin{table}
	\caption{Performance comparisons about Com\_Sco on VOC dataset}
	\label{tab:freq}
	\begin{tabular}{cccccccc}
		\toprule
		Methods &mAP & CP& CR& CF1& OP& OR& OF1\\
		\midrule
		Com\_Sco &84.5&75.4&81.4&78.3&79.9&84.3&82.0 \\
		Base &93.0&85.9&86.1&86.0&84.7&89.3&86.9 \\
		Base+ Com\_Sco&93.8&87.3&87.1&87.2&87.7&89.9&88.8\\
		\bottomrule
	\end{tabular}
\end{table}
\begin{table}
	\caption{Performance comparisons about Com\_Sco on MS-COCO dataset}
	\label{tab:freq}
	\begin{tabular}{cccccccc}
		\toprule
		Methods &mAP & CP& CR& CF1& OP& OR& OF1\\
		\midrule
		Com\_Sco &68.3&57.8&68.4&62.7&64.3&74.0&68.8 \\
		Base &82.9&83.7&72.7&77.9&84.5&76.2&80.1 \\
		Base+ Com\_Sco&84.5&86.5&73.7&79.6&87.1&77.4&82.0\\
		\bottomrule
	\end{tabular}
\end{table}
\textbf{Effectiveness of Com\_Sco }
We add the module Com\_Sco to Base+ID\_LCM+Var\_Inf for demonstrating the effectiveness of Com\_Sco module. As shown in Table 3 and Table 4, the results of comparison demonstrate its effectiveness. After obtaining the rough scores of each region on all labels, we think that score as a measure of the correlation between regions and labels can be viewed as a type of feature in label space. Since there is a correlation among labels, there should also be a correlation among regions. We view each region as a node and explore the correlation. In other words, we use the fused LCM to model a more accurate label distribution of this image instance. This module can make an effective supplement for the ResNet branch. In addition, in order to further prove the effectiveness of Com\_Sco, we do more comparative experiments. Three cases are test respectively. Firstly, we only test the region-based branch without the Var\_Inf module and the ID\_LCM module. We use the individual LCM as the adjacent matrix. Secondly, we only test the global branch without the Var\_Inf module, the ID\_LCM and the Com\_Sco module. The branch use the statistical LCM as the adjacent matrix. Thirdly, we use the fused branch without the Var\_Inf module and the the ID\_LCM module. They use the statistical LCM and the individual LCM as their adjacent matrixs respectively. The results in Table 5 and Table 6 further demonstrate its effectiveness. Although the performance of the Com\_Sco module itself is not very good, it can be used as an effective supplement to the global branch to further improve the performance.

\section{Conclusion}
How to explore label dependencies is crucial for multi-label recognition task. In order to better model the correlation among labels, we propose an IA-GCNN model which involves a global branch modeling the whole image and a region-based branch exploring dependencies among ROIs. 
We first fuse both the statistical LCM and an individual one of each image instance that is constructed by mining the label dependencies based on the features of detected ROIs, and then inject adaptive information of label-awareness into the learned features of the model for label diffusion of instance-awareness in graph convolution. Simultaneously, considering the contribution differences of ROIs to multi-label classification, we introduce variational inference to learn adaptive scaling factors for those ROIs by considering their complex distribution. Experiments on MS-COCO and VOC datasets demonstrate the superiority of our proposed model.
\bibliographystyle{ACM-Reference-Format}
\bibliography{sample-base}
\end{document}